%% file: conference.tex
\documentclass{article} 
\usepackage{iclr2021_conference,times}
\input{math_commands.tex}

\usepackage{hyperref}
\usepackage{url}
\usepackage{graphicx}
\usepackage{dirtytalk}

\title{eDarkTrends: Harnessing Social Media \\Trends in Substance use disorders for \\ Opioid Listings on Cryptomarket}
\author{Usha Lokala\textsuperscript{1}, Francois Lamy\textsuperscript{2}, Triyasha Ghosh Dastidar\textsuperscript{3}, Kaushik Roy\textsuperscript{1}, Raminta Daniulaityte\textsuperscript{4}, \\
\textbf{Srinivasan Parthasarathy\textsuperscript{5}, Amit Sheth\textsuperscript{1}}\\\\
\textsuperscript{1}Artificial Intelligence Institute, University of South Carolina, USA \\
\textsuperscript{2}Department of Society and Health, Mahidol University, Thailand\\
\textsuperscript{3}Dept. of Computer Science and Engineering, BITS Pilani, India\\
\textsuperscript{4}College of Health Solutions, Arizona State University, USA\\
\textsuperscript{5}Department of Computer Science and Engineering, Ohio State University, USA\\
amit@sc.edu, \{kaushikr, nlokala\}@email.sc.edu, f20170829@hyderabad.bits-pilani.ac.in, \\
 francois.lam@mahidol.edu, raminta.daniulaityte@asu.edu, srini@cse.ohio-state.edu
}

\iclrfinalcopy 
\begin{document}

\maketitle
\begin{abstract}

Opioid and substance misuse is rampant in the United States today, with the phenomenon known as the "opioid crisis". The relationship between substance use and mental health has been extensively studied, with one possible relationship being; substance misuse causes poor mental health. However, the lack of evidence on the relationship has resulted in opioids being largely inaccessible through legal means. This study analyzes the substance misuse posts on social media with the opioids being sold through cryptomarket listings. We use the Drug Abuse Ontology, state-of-the-art deep learning, and BERT-based models to generate sentiment and emotion for the social media posts to understand users’ perception on social media by investigating questions such as: which synthetic opioids people are optimistic, neutral, or negative about? or what kind of drugs induced fear and sorrow? or what kind of drugs people love or thankful about? or which drug people think negatively about? or which opioids cause little to no sentimental reaction. We also perform topic analysis associated with the generated sentiments and emotions to understand which topics correlate with people's responses to various drugs. Our findings can help shape policy to help isolate opioid use cases where timely intervention may be required to prevent adverse consequences, prevent overdose-related deaths, and worsen the epidemic.
\end {abstract}
\section{Introduction}
North America is facing the worst opioid epidemic in its history. This epidemic started with the mass diversion of pharmaceutical opioids (e.g., Oxycodone, Hydromorphone), resulting from the strong marketing advocacy of the potential benefits of opioids (\citealt{Lamy2020-wx}). The increase of opioid use disorder prevalence and pharmaceutical opioid-related overdose deaths resulted in a stricter distribution of pharmaceutical opioids, unintentionally leading to a dramatic increase in heroin usage among pharmaceutical opioid users (\citealt{National_Institute_on_Drug_Abuse_undated-wb}). The epidemic entered its third wave when novel synthetic opioids (e.g., fentanyl, U-47,700, carfentanil) emerged on the drug market. Several recent research and reports are pointing at the role of cryptomarkets in the distribution of emerging Novel Psychoactive Substances (NPS)  (\citealt{Aldridge2016-oj,National_Academies_of_Sciences_Engineering_and_Medicine2017-wl}). The importance of cryptomarkets has been further exacerbated by the spillover mental health and anxiety resulting from the ongoing Covid19 pandemic: recent results from the Global Drug Survey suggest that the percentage of participants who have been purchasing drugs through cryptomarkets has tripled since 2014 reaching 15 percent of the 2020 respondents (\citealt{GDS}).  
In this study, we assess social media data from active opioid users to understand what are the behaviors associated with opioid usage to identify what types of feelings are expressed. We employ deep learning models to perform sentiment and emotion analysis of social media data with the drug entities derived from cryptomarkets. We implemented LSTM, CNN, and BERT-based models for sentiment and emotion classification of social media data. Also, we performed the topic analysis using (TFIDF) to extract frequently discussed opioid-related topics in social media. 


\section{Methods}
\label{gen_inst}
\subsection{Data collection}
Concerning Dark web data, three cryptomarkets, Dream market, Tochka, and WallStreet Market, were periodically crawled in between March 2018 and January 2019. Over 70,000 opioid-related listings were collected using the dedicated crawler (\citealt{Kumar2020-is,Lamy2020-wx,Lokala2020-ay}). Raw HTML files collected were parsed and processed using a Named Entity Recognition (NER) to further extract substance names, product weight, price of the product, shipment information, availability, and administration route as shown in Table \ref{tab: tab 1}. We collected 290,458 opioid-related posts from six sub-Reddits using custom built crawlers. These posts were further processed to extract data used for social media sentiment analysis. The SubReddit corpus is spread over different drug categories such as Heroin (136,745), Kratom (77,443), Fentanyl (36,166), Oxycodone (25,890), Opium (9,675), Non-Pharmaceutical Fentanyl (2,798), Pharmaceutical Fentanyl (876), and Synthetic Heroin (865). To complete the social media emotion analysis, we also collected 21,563 posts from Twitter using Twitter API. We applied TF-IDF over unigram, bigrams, and trigrams to identify topics in each SubReddit as shown in Table \ref{tab: tab 2}.
\begin{table}[]
\begin{center}
\begin{tabular}{ |c|c| } 
 \hline
 \textbf{Property Name} & \textbf{Cryptomarket Listing Information} \\ 
 \hline
 Has Product Name & 50 Gr ***** Heroin AAA+ With Spots Free Shipping \\
 \hline
 Is Substance & Heroin \\
 \hline
 Has Class & Opiate \\
 \hline
 Has Dosage & 1.5 gram \\
 \hline
 Has Quantity & 50 gram \\
 \hline
 Has Vendor & BulkBrigade \\
 \hline
 Has Price & BTC 0.0444 \\
 \hline
 Ships To & Worldwide \\
 \hline
 Ships From & Germany \\
 \hline

\end{tabular}
\end{center}
\caption{Sample of property types identified from cryptomarket product listing}
\label{tab: tab 1}
\end{table}

\begin{table}[]
\begin{center}
\begin{tabular}{ |p{3cm}|p{12cm}| } 
 \hline
 \textbf{SubReddit} & \textbf{Topics of Interest} \\ 
 \hline
 Opiates Recovery & Cold turkey withdrawal, cravings, anxiety, rehab, depression, sobriety, Loperamide, Benzo, Subutex, quitting, Vivitrol, Imodium, Naltrexone \\
 \hline
 Opiates & Codeine, Hydrocodone, Oxymorphone, Dilaudid, hydromorphone, Opana, Oxycontin, Acetaminophen, Gabapentin, benzos, Roxicodone \\
 \hline
 Suboxone & Buprenorphine, Subutex, Agonist, Clonidine, Tramadol, Hydrocodone, Dilaudid, Vicodin, Sublocade, Percocet, Phenibut, Klonopin, Valium \\
 \hline
 Heroin & Dope, Opium, Opiates, Crack, Diacetylmorphine, China White, codeine, acetaminophen \\
 \hline
 Drug Nerds & Methadone, Alkaloids, Mitragynine, Benzos, Poppy, Buprenorphine, Antagonist, Gabapentin, Naloxone, Amphetamine, Hydrocodone \\
 \hline
 Research Chemicals & Benzos, Psychoactive, Psychedelic, Kratom, Pyrovalerone, Quaalude, Oxycodone, Morphine, Xanax, Tramadol, Cocaine, Methadone, Ketamine, Gabapentin, Amphetamine, Hydromorphone \\
 \hline
\end{tabular}
\end{center}
\caption{Sample of Topics identified from different subreddits}
\label{tab: tab 2}
\end{table}
\subsection{Named Entity Recognition}
We used a pre-trained NER deep learning (NER DL) bidirectional LSTM-CNN approach (\citealt{Chiu2016-dm}) on crypto market data to identify drug entities that use a hybrid bidirectional LSTM and CNN architecture, eliminating the need for most feature engineering. The entities are then matched to a superclass using Drug Abuse Ontology (DAO) (\citealt{Cameron2013-ho,Lokala2020-ay}) that acts as a domain-specific resource with all superclasses related to the entities. We identified 90 drug entities, which we then broadly classified into eight categories by mapping each entity to a super drug class in DAO. The eight broad categories considered are Heroin, Synthetic Heroin, Pharmaceutical Fentanyl, Non-Pharmaceutical Fentanyl, Fentanyl, Oxycodone, Kratom, and Opium.

\subsection{Sentiment Analysis}
We classified SubReddit posts as Positive, Negative, and Neutral categories for sentiment analysis. We implemented Textblob to generate sentiment for each SubReddit post. The TextBlob generated labels are used as a training set to implement SOTA DL algorithms like CNN, LSTM, and language model  BERT. The highest accuracy achieved is 93.6 with the LSTM model. We report the stats of Sentiment labels for SubReddit posts obtained from sampling 800 random data points from each drug category are reported in Table \ref{tab: tab 3}.
\begin{table}[!h]
\begin{center}
\begin{tabular}{ |p{2cm}|p{2cm}|p{2cm}|p{2cm}|p{5.5cm}| } 
 \hline
 \textbf{Drug} & \textbf{Positive} & \textbf{Negative} & \textbf{Neutral} & \textbf{Top 3 Emotions in the order found}\\ 
 \hline
Opium & 481 & 218 & 101 & Sadness, Love, Joy \\
 \hline
 Oxycodone & 460 & 245 & 95 & Sadness, Fear, Thankfulness\\
 \hline
 Kratom & 459 & 231 & 110 & Love, Sadness, Fear \\
 \hline
 Fentanyl & 467 & 274 & 59 & Sadness, Love, Fear/Thankfulness \\
 \hline
 Heroin & 455 & 255 & 90 & Sadness, Joy, Thankfulness\\
 \hline
 Synthetic Heroin & 500 & 240 & 60 & Sadness, Fear, Thankfulness\\
 \hline
 Pharmaceutical Fentanyl & 570 & 197 & 33 & Sadness, Love, Joy/Thankfulness\\
 \hline
 SNon-Pharmaceutical Fentanyl & 502 & 264 & 34 & Sadness, Love, Thankfulness\\
 \hline

\end{tabular}
\end{center}
\caption{ Sentiment stats(Number of Posts) after sampling 800 random points for each drug category identified from 6 Subreddits and Top emotions identified for each drug from Twitter}
\label{tab: tab 3}
\end{table}
\subsection{Emotion Analysis}
We did not choose to work on SubReddit data as we do not have self-tagged emotions in posts on SubReddit. Therefore, for Emotion analysis, we decided to crawl Twitter, where emotions are present as hashtags. We limited our crawl to 7 kinds of emotions, as stated in work done by Wang et al (\citealt{Wang2012-bn}). We extracted labeled training data by crawling tweets with each emotion’s hashtags: Joy, Sadness, Anger, Love, Fear, Thankfulness, and Surprise. We then trained deep learning models LSTM, CNN, fine-tuned BERT-based models, and generated emotion labels for drug-based Twitter data. The highest accuracy achieved is 91.2 for the LSTM model.  
\section{Results}
\begin{figure*}[!h]
    \centering
    \includegraphics[width=\textwidth]{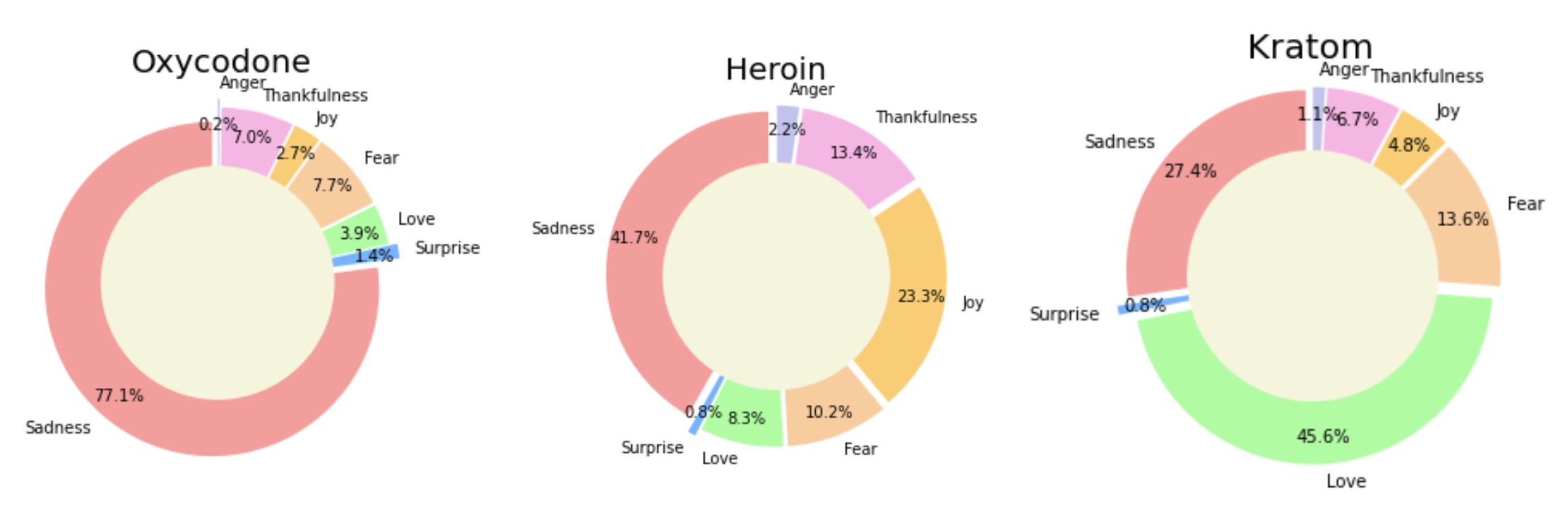}
    \caption{Comparison for drugs Kratom, Heroin, and Oxycodone by seven emotions: Joy,  Sadness, Anger, Love, Fear, Thankfulness, and Surprise}
    \label{fig:fig1}
\end{figure*}

Kratom, Heroin, Fentanyl, Morphine, Cocaine, Methadone, Suboxone, and Oxycodone are the commonly discussed drugs across six subreddits. In Table 2. For example, consider Research chemicals (RC); it is interesting to find that more posts talk about Pyrovalerone, a psychoactive drug with stimulant effects. Another term found is ‘Quaalude,’ a brand name for ‘Methaqualone,’ a sedative and hypnotic medication. The RC subreddit mostly discusses psychoactive and psychedelic drugs, while DrugNerds discusses Alkaloids (\citealt{Kaserer2020-dk}). Interestingly, DrugNerds talks about Naloxone, which can treat Opioid overdose. Dope is a slang term for Heroin identified in Heroin Subreddit. Several brand names of medications for anxiety, pain, seizures, insomnia, and sedatives are discussed in the Suboxone subreddit. Gabapentin is the typical seizure and pain medication discussed among most of the subreddits. Opiates Recovery is more about the withdrawal symptoms and mental health disorders, for example, ‘cold turkey.’ The ‘cold turkey’ used in the context of substance misuse is quitting substance abruptly, which carries significant risks if the drug you are discontinuing is a benzodiazepine or opiate (\citealt{Just2016-bk,Landry1992-ge}). The results show that we can derive slang terms, brand names, novel drugs, mental health symptoms, and medications from social media. 
From the results in Table \ref{tab: tab 3}, It is found that the highest positive sentiment is found in Pharmaceutical Fentanyl, the highest negative sentiment for Fentanyl, and the highest neutral opinion for Kratom. The emotion 'Love' is detected the top one for Kratom as people use it for self medication. The emotions among Twitter data for Fentanyl, Heroin, Oxycodone are visualized in Figure \ref{fig:fig1} with seven emotions: Joy,  Sadness, Anger, Love, Fear, Thankfulness, and Surprise. 
The top three emotions for each drug are presented in Table \ref{tab: tab 3}. The results for three deep learning approaches for sentiment analysis and emotion analysis: F-measure, Precision, and Recall are reported in Table \ref{tab: tab 4}.
\begin{table}[!h]
\begin{center}
\begin{tabular}{ |c|c|c|c| } 
 \hline
 \textbf{Sentiment Model} & \textbf{P} & \textbf{R} & \textbf{F1} \\ 
 \hline
LSTM & 0.937 & 0.936 & 0.936 \\
 \hline
 CNN & 0.886 & 0.884 & 0.885 \\
 \hline
 BERT & 0.884 & 0.883 & 0.825 \\
 \hline
 \textbf{Emotion Model} & \textbf{P} & \textbf{R} & \textbf{F1}  \\
 \hline
 LSTM & 0.912 & 0.912 & 0.912 \\
 \hline
 CNN & 0.884 & 0.883 & 0.883 \\
 \hline
 BERT & 0.781 & 0.781 & 0.779 \\
 \hline

\end{tabular}
\end{center}
\caption{ Performance Comparison of Deep Learning models for Sentiment and Emotion Analysis}
\label{tab: tab 4}
\end{table}
\section{Discussion}
Crawling cryptomarkets poses a significant challenge to apply data science and machine learning to study the opioid epidemic due to the restricted crawling process (\citealt{Kumar2020-is,Lamy2020-wx,Lokala2020-ay}). To identify the best strategies to reduce opioid misuse, a better understanding of cryptomarket drug sales that impact consumption and how it reflects social media discussions is needed (\citealt{Kamdar2019-lm}). Since our social media data is based on eight broad category drugs, we hope to further refine our categories by consulting with a domain expert.
Further, we have identified the processes for future research. We plan to expand this work to extract mental health symptoms from the drug-related social media data to connect the association between drugs and mental health problems, for example, the association between cannabis and depression (\citealt{Roy2021-yw,Yadav2021-zx}). We also plan to build an Opioid Drug Social Media Knowledge graph with all these different data points (Drug, Sentiment, Emotion, mental health symptom, location) and compare it against the work on ‘Knowledge Graph-based Approach For Exploring The U.S. Opioid Epidemic’ (\citealt{Kamdar2019-lm}). Potential areas of application would be identifying risk factors regarding addiction and mental health from subreddit data (\citealt{Gaur2018-gj}), identifying drug trends based on location, and predicting Opioid Overdoses. We would also like to rely on DEA Drug Seizures to include in our preliminary data collection process to be aware of related social media discussions.



\subsubsection*{Acknowledgments}
We acknowledge partial support from the National Institute on Drug Abuse (NIDA) Grant Number: R21DA04451: \say {eDarkTrends:Monitoring Cryptomarkets to Identify Emerging Trends of Illicit Synthetic Opioids Use} and NSF Award Number: 1761969: \say {Spokes: MEDIUM: MIDWEST: Collaborative: Community-Driven Data Engineering for Substance Abuse Prevention in the Rural Midwest}. All findings and opinions are of authors and not sponsors.

\bibliography{conference}
\bibliographystyle{conference}


\end{document}

%% file: math_commands.tex
\usepackage{amsmath,amsfonts,bm}

\def\eqref#1{equation~\ref{#1}}

\def\1{\bm{1}}

\DeclareMathAlphabet{\mathsfit}{\encodingdefault}{\sfdefault}{m}{sl}
\SetMathAlphabet{\mathsfit}{bold}{\encodingdefault}{\sfdefault}{bx}{n}

